\documentclass[letterpaper, 10 pt, conference]{ieeeconf}  

\IEEEoverridecommandlockouts                              

\usepackage{graphics} 
\usepackage{graphicx}
\usepackage{amsmath}
\usepackage{color}
\usepackage[ruled,linesnumbered]{algorithm2e}
\usepackage{booktabs}
\usepackage{multirow} 
\usepackage{bbm}
\usepackage{amssymb}
\usepackage{tabularx}
\usepackage{graphicx}
\usepackage{epstopdf}
\usepackage{todonotes}
\usepackage{multirow}
\usepackage{bm}
\usepackage{gensymb}
\usepackage{array,multirow}
\usepackage{array}
\usepackage{makecell}

\title{\LARGE \bf
SRM: An Efficient Framework for Autonomous Robotic Exploration in Indoor Environments
}

\author{Chaoqun Wang$^{1}$, Delong Zhu$^{1}$, Teng Li$^{2}$, Max Q.-H. Meng$^{1*}$, and Clarence De. Silva$^{2}$ 
\thanks{$^{*}$Corresponding Author}
\thanks{$^{1}$Chaoqun Wang, Delong Zhu and Max. Q. H. Meng are with The Chinese University of Hong Kong, China.{\tt\small \{cqwang, dlzhu, qhmeng\}@ee.cuhk.edu.hk} }
\thanks{$^{2}$Teng Li  and Clarence W. de Silva are with The University of British Columbia, Vancouver, BC, Canada.   {\tt\small \{tengli,desilva\}@mech.ubc.ca}
}
}

\begin{document}

\maketitle
\thispagestyle{empty}
\pagestyle{empty}

\begin{abstract}
In this paper, we propose an integrated framework for the autonomous robotic exploration in indoor environments. Specially, we present a hybrid map, named Semantic Road Map (SRM), to represent the topological structure of the explored environment and facilitate decision-making in the exploration. The SRM is built incrementally along with the exploration process. It is a graph structure with collision-free nodes and edges that are generated within the sensor coverage. Moreover, each node has a semantic label and the expected information gain at that location. Based on the concise SRM, we present a novel and effective decision-making model to determine the next-best-target (NBT) during the exploration. The model concerns the semantic information, the information gain, and the path cost to the target location. We use the nodes of SRM to represent the candidate targets, which enables the target evaluation to be performed directly on the SRM. With the SRM, both the information gain of a node and the path cost to the node can be obtained efficiently. Besides, we adopt the cross-entropy method to optimize the path to make it more informative. We conduct experimental studies in both simulated and real-world environments, which demonstrate the effectiveness of the proposed method. 



\end{abstract}

\section{INTRODUCTION}
Autonomous exploration has an increasing number of applications such as disaster relief, search and rescue, and environment monitoring. It is a fundamental capability for the mobile robot to gather information of an unknown environment. The primary goal is to autonomously gather required information with limited time or resource budget, which poses great challenges for exploration strategies. 


Various past methodologies have been proposed to tackle the  autonomous exploration problem, among which the most investigated approach is the nearest frontier method \cite{yamauchi1997frontier}. The frontier in this method is defined as the boundary between the known and unknown regions on the incomplete map. In this scheme, the robot always pursuits the nearest frontier to explore. Besides the frontier-based method, the information gain is also frequently adopted in the context of autonomous exploration \cite{stachniss2005information}, where the NBT is determined by evaluating the amount of information gain at the target areas. Typically, these approaches use the metric map to represent the environment and extract the required information, which can be time-consuming in large-scale environments.

In contrast with the majority of frontier or information  based methods, little effort has been devoted to developing special-designed path planning methods for the exploration. However, an efficient path planner is a prerequisite for the target evaluation involving path cost.  Sampling-based path planning method \cite{lavalle1998rapidly} is a broadly adopted approach in the exploration. It is more efficient with respect to the large-scale or high-dimensional environments. However, these planning methods are not historical-aware. A path will be replanned using these methods on the whole map when the robot needs to backtrack to some places that is previously visited without further exploitation. Besides, the previous planning methods pay more attention to the path cost without considering the information collected along the path.


\begin{figure}[t]
\centering
\includegraphics[width=8cm,height=5cm]{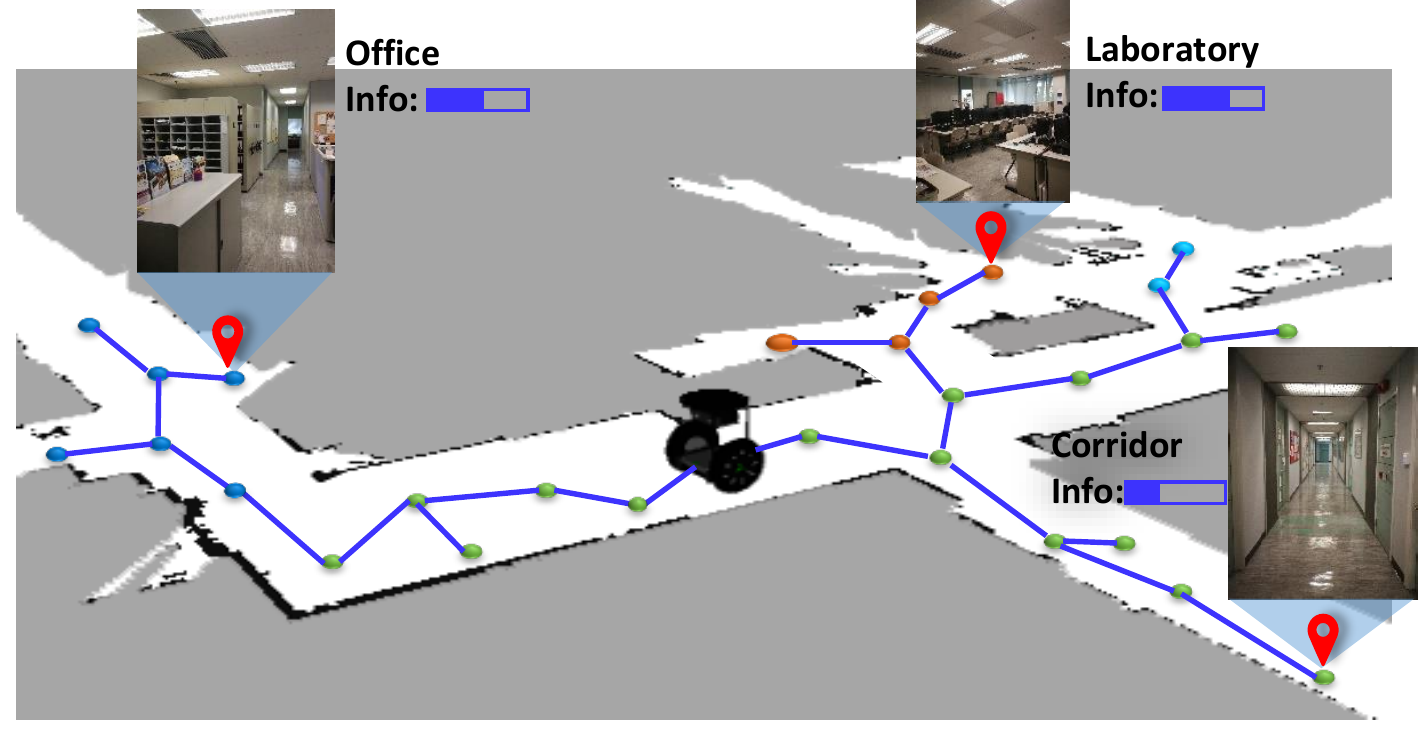}
\caption{SRM during the exploration. The blue lines and the colourful points represent the edges and the nodes of the graph, respectively. There is a semantic label and the information gain for each node.}
\label{fig:startPic}
\end{figure}


In this paper, we propose to use a novel hybrid map, the SRM, to guide the exploration. This topological map is a graph structure that is built incrementally over the course of exploration. As shown in Fig. \ref{fig:startPic}, the node on the graph structure has two attributes: the semantic label and the information gain at the corresponding location. The NBT selection is performed directly on this topological map. Based on the SRM, we propose a hierarchy decision-making strategy, where the high-level decision is made according to the semantic information, and the low-level decision making is concerned with the path cost and the information gain. Taking advantage of such a process, not only can we get the information gain efficiently comparing with computing on the whole metric map, but we can also query a path cost directly on the graph structure. Furthermore, the queried path is optimized using the cross-entropy method.  Hence the robot can get the most information along the generated trajectory. The experimental results demonstrate that our method can explore the unknown environment more efficiently than the conventional methods.

The remaining of this paper is organized as follows. In Section. \ref{sec:related work}, we introduce some related work about autonomous exploration. Then in Section. \ref{sec:preliminaries}, we give some preliminaries about our problem. After that, we  introduce our methodology in Section. \ref{sec:methodology}. Our experiments results are presented in Section. \ref{sec:experiments and results}. Finally, we give the conclusion and future work in Section. \ref{sec:conclusion and future work}.

\section{Related work}
\label{sec:related work}
%


A large number of related approaches are developed to cope with different aspects of the autonomous exploration problem \cite{mallios2016toward}.  In this section, we mainly focus on three tightly coupled aspects that our framework is built upon in the autonomous exploration. To start with, we give an overview of the map representation in robotic navigation. Then we describe different decision-making strategies in the exploration process. Finally, we review the literature on informative path planning during the exploration. 

\subsection{Map representation}
The study of map representation in robot tasks is one of the most active areas in robotics research. Generally speaking, the map can be divide into four groups: the metric map, the topological map, the semantic map, and the hybrid map \cite{thrun2002robotic}. Various approaches have been proposed to enable the robot navigation with different kinds of maps \cite{konolige2011navigation}, but few studies have been focused on the map representation during the exploration. 
Choset \textit{et al.} \cite{choset2000sensor} proposed a sensor based hierarchically generated Voronoi graph to represent the environment. One can query a path efficiently and completely on the graph during the exploration. Rezanejad \textit{et al.} \cite{rezanejad2015robust} suggest to use the topological map generated by the flux skeleton based method on the occupancy grid map, where the skeleton is generated online by utilizing image-based method instead of constructing incrementally with the exploration.

\subsection{Guidance of exploration}
NBT evaluation is another critical issue in  autonomous exploration. Two  frequently used metrics for determining the NBT are the nearest frontier and the information gain of the candidate targets \cite{keidar2014efficient}\cite{jadidi2018gaussian}. Shen \textit{et al.} \cite{shen2012autonomous} improve the exploration efficiency through evaluating the NBT according to a stochastic differential equation-based method. Charrow \textit{et al.}  proposed to use the information-theoretic objective function to guide the exploration.  More recently, Umari \textit{et al.} \cite{umari2017autonomous} present an exploration method based on rapidly exploring random tree (RRT). The frontier regions can be detected efficiently through sampling based method. Besides, some novel metrics are also proposed for the evaluation of NBT. Dornhege \textit{et al.} \cite{dornhege2013frontier} present a frontier-void-based method that uses both the frontier and the unexplored 3D volumes to guide the exploration. Visser \textit{et al.} \cite{visser2008balancing} propose to balance the path cost and the information gain during the exploration.

In recent years, there has been growing interests in making the autonomous exploration more intelligent. O$ß$wald \textit{et al.} \cite{osswald2016speeding}  present an exploration strategy by exploiting the structure of the environment. Semantic information has been widely used for the robot navigation as well \cite{stachniss2009multi}.  With semantic reasoning, a robot can achieve human-like performance.  Moreover, some learning-based methods toward autonomous exploration have been proposed  along with the developments in deep learning approaches \cite{zhu2018rlsuper}. 

\subsection{Informative path planning}

Path planning is of great importance during the exploration but most work mainly pays attention to the motion constraints without considering the information gathered along the path. Heng \textit{et al.} \cite{heng2015efficient} propose an algorithm for visual exploration considering both the exploration and the coverage. The robot is pruning to select a path with more information. Davis \textit{et al.} \cite{davis2016c} present a coverage aware path planning method. This approach plans a path from a start point to the goal while maximizing the user defined regions.  Tan \textit{et al.} \cite{Tan:icra2017} propose an information gathering method based on cross-entropy. The trajectory is optimized using the cross-entropy method to maximize the objective function in information gathering.

\section{PRELIMINARIES}
\label{sec:preliminaries}

Our main goal is to build a precise metric map of an unknown environment with as less path cost and exploration time as possible.  We use the occupancy grid map to model the environment. The map, denoted as $\mathcal{M}_o \subset \mathbb{R}^2$, encodes the probability of being occupied for every grid. Suppose there are $m$ candidate regions $\{x_1,x_2,...,x_m\}$ to be explored, the information gain $I_i$ of the candidate region $x_i$ is obtained using $\mathcal{M}_o$. The entropy over the occupancy grid map $\mathcal{M}_o$ is denoted as: 

\begin{equation}
\label{information gain}
H(\mathcal{M}_o)=-\sum_i \sum_j p(c_{i,j})\log p(c_{i,j}),
\end{equation}
where $c_{i,j} \in \mathcal{M}_o$ is the grid cell of the map at location $\{i,j\}$.
We use mutual information to represent the information gain for a sensing location $x_i$, which is defined as: 
\begin{equation}
\label{eq:mutual information}
I(\mathcal{M}_o;x_i)=H(\mathcal{M}_o)-H(\mathcal{M}_o|x_i)
\end{equation}

We propose to use the SRM for the decision-making process over the course of exploration. The SRM, denoted as  $\mathcal{G}=\{V,E\}$, is the topological map of the explored environment. It is a graph structure with $V$ representing the vertices and with $E$ represents the edges that connect the vertices. For every $v_i \in V$ on the graph $\mathcal{G}$, it has two attributes: the semantic label $S$ and the information gain $I_i$ during the exploration. A semantic mapping process is maintained and the semantic label is continually attached to the nodes of the graph. The nodes on the graph are used to indicate the candidate regions to be explored, i.e. $v_i \equiv x_i$.
The path $\hat{\mathcal{P}}$ from the robot current location $x_o$ to the target area $x_i$ can be queried efficiently on the graph $\mathcal{G}$. Then the path cost $\ell(\hat{\mathcal{P}})$ is calculated by:
\begin{equation}
\ell(\hat{\mathcal{P}})= \mathcal{L}(\hat{\mathcal{P}},\mathcal{M}_o).
\end{equation}
The path $\hat{\mathcal{P}}$ will be optimized with respect to $\mathcal{M}_o $ utilizing the function $\mathcal{L}(\cdot)$, which is the trajectory optimization method based on cross-entropy. 

\section{Methodology}
\label{sec:methodology}
\subsection{Method Overview}
\begin{figure}[t]
\centering
\includegraphics[width=0.5\textwidth]{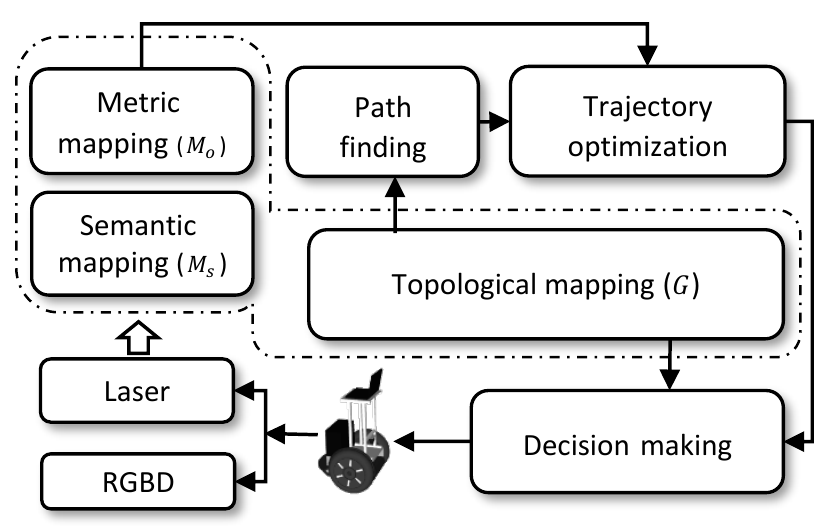}
\caption{Overall framework of our proposed SRM method. Details can be found in the text.}
\label{fig:framework}
\end{figure}

To build a precise metric map of the environment, we build a topological map incrementally during the course of exploration. As Fig. \ref{fig:framework} shows, the mapping process maintains three maps: the occupancy grid map $\mathcal{M}_o$, the semantic map $\mathcal{M}_s$, and the graph structure $\mathcal{G}$. The node on the graph $\mathcal{G}$ contains the semantic label provided by $\mathcal{M}_s$ and the information gain queried on $\mathcal{M}_o$.  Robot can find a path efficiently on the graph $\mathcal{G}$, which is then optimized by the cross-entropy based method. The determination of NBT involves not only the path cost and the information gain, but also the high-level semantic information. All these evaluation metrics are combined in the decision-making process to guide the robotic exploration. 


\subsection{Semantic Road Map}
\subsubsection{Tree construction}

\begin{figure}[t]
\centering
\includegraphics[width=6cm,height=4cm]{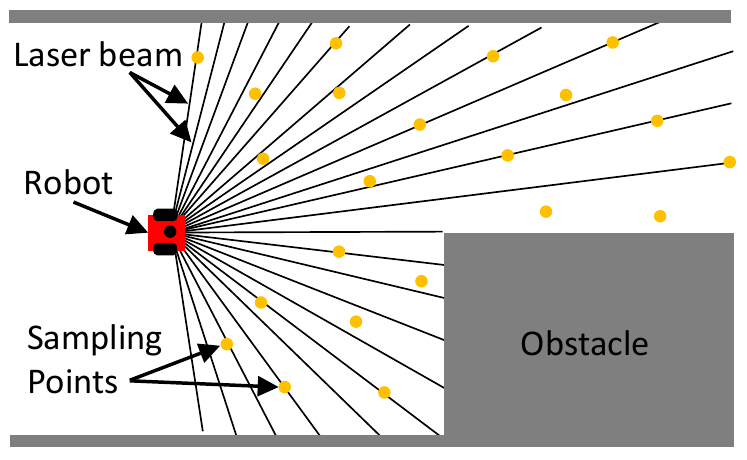}
\caption{Sampling under a sensor coverage. The yellow points represent the sampled points under the current sensor scope.}
\label{fig:new sample}
\end{figure}

The graph structure $\mathcal{G}=\{ V,E \} $ is built incrementally along with the exploration process. Initially, there is only one original node root at the starting point where the robot starts to explore the environment, i.e. $V= \{ v_{root} \}, E=\emptyset$. The pseudo code of building the graph structure is shown in Alg. \ref{alg: graph generation}.  We propose to use the sampling points under the sensor scope as the candidate vertices for building the graph. As shown in Fig. \ref{fig:new sample}, the laser beams indicate the current sensor scope. Function $Random(\cdot)$ takes as input the sensor scope and generate sampling points $V_{rand}$, as indicated by the yellow points in Fig. \ref{fig:new sample}. For a random point $v_r \in V_{rand}$, it will find the nearest vertex $\hat{v}$ on graph $\mathcal{G}$ according to:
\begin{equation}
\hat{v}= \operatorname*{argmin}_{v_i \in (V\setminus V')} ||v_i-v_{r}||, V \in \mathcal{G},
\end{equation}
where $V'$ is the set of the vertex on the graph and the connection from $v_i \in V'$ to $v_r$ is not collision free. If $V \setminus V' = \emptyset$, then $v_r$ is not a feasible random point since $v_i$ does not exist. Function $Random(\cdot)$ will be called several times before a feasible sample point is obtained.  Then, the feasible point $v_r$ connect the nearest vertex in the graph $\mathcal{G}$. The random point $v_r$ and the edge $e$ that connects $\hat{v}$ and $v_r$ are added to the existing graph. Furthermore, each effective vertex $v_r$ is attached with a semantic label which corresponds to the room type detected by the semantic mapping pipeline. However, if $l=||\hat{v}-v_r||$ is less than a threshold, then the random point will be dropped and a new iteration will start.  Line 10-22 in Alg. \ref{alg: graph generation} describe the details of adding a vertex to the graph. The algorithm runs online along with the exploration process.


\begin{algorithm} [t]
\KwIn{laserMsg, rgbdMsg}
\KwOut{semantic road map $\mathcal{G}$}
\SetAlgoLined
 \While{existing unexplored area}{
$\mathcal{M} \leftarrow metricMapping(laserMsg)$\;
$V.pushback(start_{node})$ \;
//generate candidate vertices \;
\For{$n_{th}$ ray of LaserMsg}{
   $v_{rand}= Random[0, laserMsg.dist(n_{th})]$ \;
   $V_{rand}.pushback(v_{rand})$ \;
}
//check validity of edges \;
\For{$i=1 \ to \  Size(V_{rand})$}{
$rank(V_{rand}[i],V,ascend)$\;
\For{$ j = 1 \ to \ Size(V)$}{
   $ edge=Line(V_{rand}[i],V[j]) $  \;
   \If{CheckValidity(edge,$\mathcal{M}$)}{
   $E.pushback(edge) $\;
   $V_{rand}[i].label = Label(rgbdMsg,V_{rand}[i])$\;
   $V.pushback(V_{rand}[i]) $\;
   break; // to simplify the graph
   }
}
}
}
\caption{Semantic Topological Map Generation}
\label{alg: graph generation}
\end{algorithm}

We use A* to search a global path $\hat{\mathcal{P}}$ on the existing graph $\mathcal{G}$. Since we use the node on the graph to represent the target, it is simple and straightforward to get a sparse path on the graph. 
Some  edges on the graph may become infeasible along with the exploration process. The reason for this is that the collision checking is performed on the incomplete occupancy grid map. The edges that lie in the unknown area in the map is assumed to be collision-free, but sometimes there could be collision in the unknown area. 
In our proposed method, we check the feasibility of the path only if a path is generated. If the generated path is not feasible, i.e. one or more edges for constructing the graph are not collision free, we delete those edges on the graph and restart the path-finding process.

\subsubsection{Efficient frontier detection}

\begin{figure}[t]
\centering
\includegraphics[width=7.5cm,height=4cm]{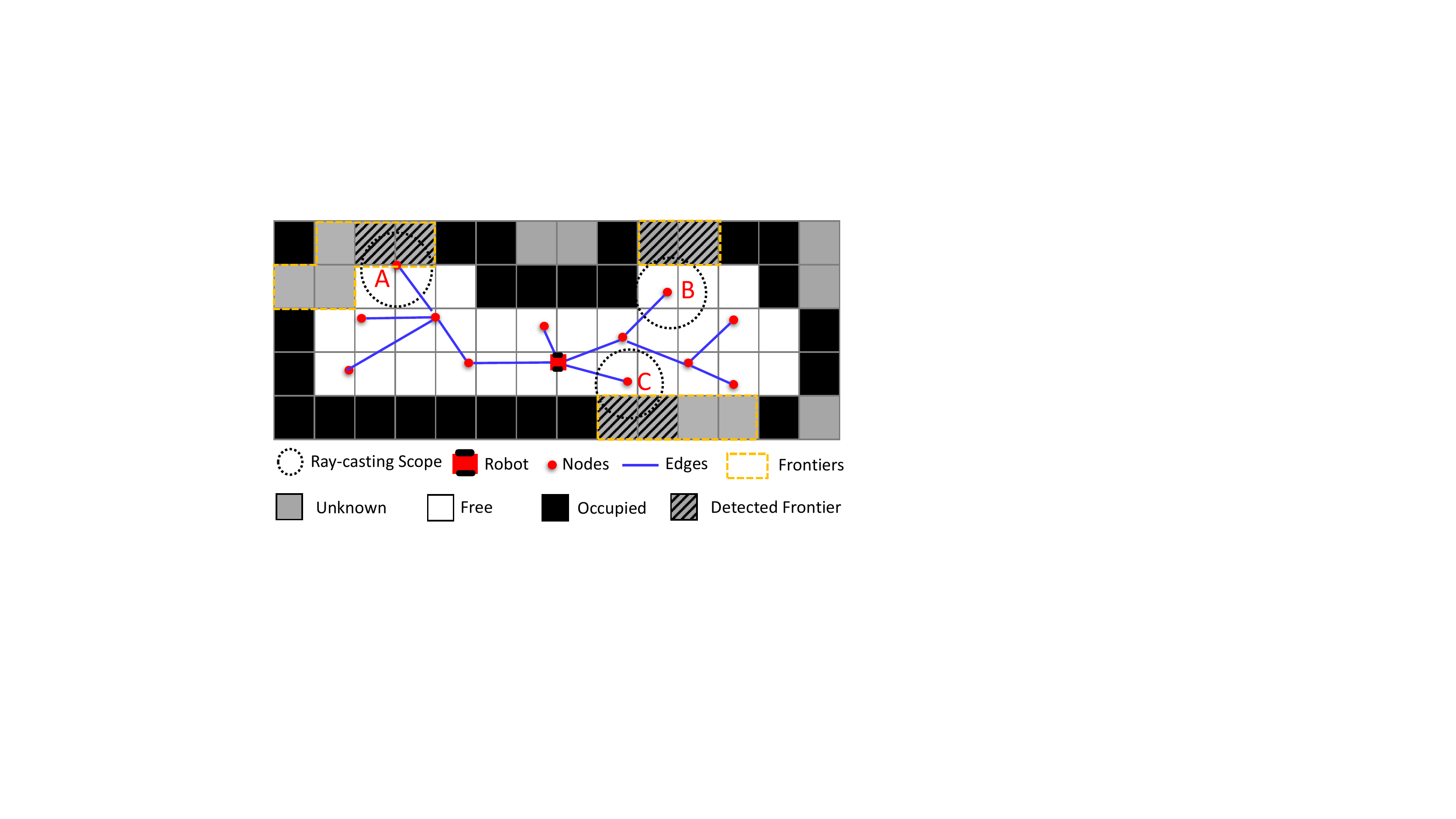}
\caption{Efficient frontier detection. The red points indicate the nodes and the blue lines represent the edges on the graph. The dotted circle represents the scope of the ray-casting method.}
\label{fig:frontierDetect}
\end{figure}

We propose an efficient frontier detection method based on the observation that the frontier cells are always appears grouply, as Fig. \ref{fig:frontierDetect} shows.  When one frontier cell $f$ is found, then the adjacent cell may also the frontier cell.  A frontier clusters can be obtained efficiently through exploring the surrounding area of a frontier $f$. To get the the cell $f$, the nodes on $\mathcal{G}$ are used for the evaluation. The graph $\mathcal{G}$ exists and extends towards the collision free area, while the frontier cells lie between the collision free area and the unknown regions. Hence the frontier point can be found near the nodes on the graph $\mathcal{G}$ inherently. We exploit the vertices on the graph $\mathcal{G}$ to find the frontiers over the course of exploration.

\begin{algorithm}[t]
\SetAlgoLined
\KwIn{$\mathcal{M}$, $v \in　\{ V \setminus V_{closed}　\}$}
\KwOut{frontier set $\mathcal{F}$}
$ P_r = rayCasting(v)$\;
\For{$m=1 \ to \ Size (P_r)$}{
$P_{cand}={P_r(m)}$\;
\While{notEmpty($P_{cand}$)}{
\For{$n=1 \ to \ Size (P_{cand})$}{
\If{$isFrontier(P_{cand}(n))$}{
$[P_n,F_i] = findFrontierNeighbor(P_{cand}(n),\mathcal{M})$;
$P_{total}.pushback(P_n)$\;
$\mathcal{F}.pushback(F_i)$\;
}
}
$P_{cand}=P_{total}$\;
}
}
\If{$isEmpty(\mathcal{F})$}{$V_{closed}.pushback(v)$\;}
\caption{Fast Frontier Detection.}
\label{alg: fast frontier detection}
\end{algorithm}

As shown in Fig. \ref{fig:frontierDetect},  The shaded square indicates the frontier cells around the nodes detected by the ray-casting method. Two frontier cells are detected around node \textbf{B} on the $\mathcal{G}$. Note that there is another frontier cell that is not detected by the raycasting method. This cell can be found efficiently through checking the neighbors of detected frontier cells. Hence all the frontier cells can be obtained through this method. Unlike the conventional method that checks all the cells to get the frontiers. our approach is more efficient because we only need to check the cells around the nodes on the graph. To further improve the efficiency of our method, we divide the nodes on the graph $\mathcal{G}$ into two groups: the closed nodes $V_{closed}$ and working nodes $V_{open}$. The closed nodes are those no frontiers can be detected with. If there is no frontier detected around a nodes using the ray-casting method, then this nodes is closed. We will not check these nodes when performing frontier detection. This prune method is a simplify that can further improve the efficiency since the ray-casting method can also be time consuming. 

The proposed frontier detection method is shown in Alg. \ref{alg: fast frontier detection}.
A cluster of points $P_r$ can be found through function $rayCasting$ around node $v$. For every point in the set $P_r$, if it is a frontier cell, then we mark this point on the map and check its neighbor. If the neighbors are also frontiers, this process repeats until all the frontiers in the neighborhood are found. 
Line 2-13 in Alg. \ref{alg: fast frontier detection} describes the frontier detection process. 
This iteration ends when there is no frontier detected. This means that there is no frontier regions around the vertex $v$, we will put this vertex into closed set and will not check it in the following iterations.

We use the amount of frontiers to quantify the expected information gain $I$ around a node. With the proposed frontier detection method, we can obtain the information efficiently.

 
\subsection{Informative path planning}

The path $\hat{\mathcal{P}}$ connecting the start and the goal region contains the vertice $\{v_1,v_2,...,v_n\}$ and the edges $\{ e_1,e_2,..., e_{n-1}\}$. This coherent path is not optimal since the vertice are generated randomly without considering any constraints. In the proposed framework, the robot is required to gather more information when it moves from one point to another. The path should not only be with less path cost but it should also be more informative. The goal is to compute a trajectory that satisfies:
\begin{equation}
\begin{aligned}
& \mathcal{P}=\mathop{\arg\min}_{\mathcal{P}} C(\mathcal{P}) \\ 
& s.t. \ \  F(\mathcal{P,\ell(\hat{\mathcal{P}})}) \geq 0,
\end{aligned}
\end{equation}
where  $C(\mathcal{P})$ is the reward function over the path. We consider both the path cost and the information gain $I$ along the path. Thus we define the reward function by: $C=-I(\mathcal{P},t)e^{-\lambda \ell(\mathcal{P})}$, as modeled in \cite{gonzalez2002navigation}, where $\lambda$ is a positive constant which determines trade-off between the exploration and exploitation, $\ell(\mathcal{P})$ is the trajectory length of the path, and $I(\mathcal{P},t)$ is the information gain along the path defined by Equation. \ref{eq:mutual information}. $F(\mathcal{P,\ell(\hat{\mathcal{P}})}) \geq 0$ guarantees the path $\mathcal{P}$ is collision-free and satisfies the motion constrains of the robot. Besides that, this function gives the upper bound of the path cost of the final path. $\ell(\hat{\mathcal{P}})$ is the path length of the sparse path $\hat{\mathcal{P}}$. The optimized path length, $\mathcal{P}_{opt}$ should satisfy the constraint that: $\ell(\mathcal{P}_{opt}) \leq \ell(\hat{\mathcal{P}})$. This constraint can helps when the optimization of the path does no converge. Moreover, the information gain along a path is a function about time $t$, which means that the sensor scan at time t $S_t$ should consider the effect of its previous scan $S_{t-1}$. This can help the robot avoid the local minimum in information rich area using our proposed objective function.

The objective function is optimized using the cross-entropy method. In what follows we will mainly focus on how to optimize our problem using the cross-entropy method. We refer the interested reader to \cite{Tan:icra2017}\cite{kobilarov2012cross} for more details. Our development follows closely \cite{Tan:icra2017} using notation adapted to our setting. The path $\mathcal{P}$ is parameterized by the function $\mathcal{P} = \phi (z)$, where $z \in \mathcal{Z}$ is the parameter space. The cost function can then be defined by:

\begin{equation}
\begin{aligned}
& z=\mathop{\arg\min}_{z} C(\phi(z)) \\ 
& s.t. \ \  F({\phi(z)},\ell(\hat{\mathcal{P}}))) \geq 0,
\end{aligned}
\label{eq:goal function}
\end{equation}
 
This optimization in Equation. \ref{eq:goal function} can be solved by the cross-entropy method. Denote $r^*$ as the optimal reward, i.e.
\begin{equation}
r^*=\min J(z),
\label{eq:optimal}
\end{equation}
where $J(z)=\int^{T}_0 C(\phi(z))$ is the accumulated reward.
The parameter space $\mathcal{Z}$ has a pdf $p(\theta)$. Here in our case the pdf is a mixture of Gaussian, where the number of Gaussian components, n, is determined by the vertices on the original path $\hat{\mathcal{P}}$. Equation. \ref{eq:optimal}  can be formulated as:
\begin{equation}
\zeta = \mathbb{P}_{\theta}(J(Z)\leq r)= \mathbb{E}_{\theta}[I_{\{J(Z)\leq r \}}]
\end{equation}

The optimal pairs$(r,\theta)$ can be updated by:

\begin{enumerate}
\item Sample with current pdf $p(\theta)$ to get $\{z_1,z_2,...,z_m\}$, determine $R=\{\hat{r_1},\hat{r_2},...,\hat{r_m}\}$ by Equation. \ref{eq:goal function}
\item Select the elite set $R^*$, compute the $\rho$-th quantile $r_j$
\item Update $\theta$ by:
\begin{equation}
\theta=\mathop{\arg\min}_{\theta}\frac{1}{|\varepsilon|}\sum_{Z_k \in \varepsilon} ln p(Z_k;\theta)
\end{equation}
\end{enumerate}

This iteration will end until a certain termination condition is fulfilled. Then we get an optimized trajectory  $\mathcal{P}_{opt}$, which is more informative with less path cost as well. 

\subsection{Decision making}

\begin{figure}[t]
\centering
\includegraphics[width=6cm,height=4cm]{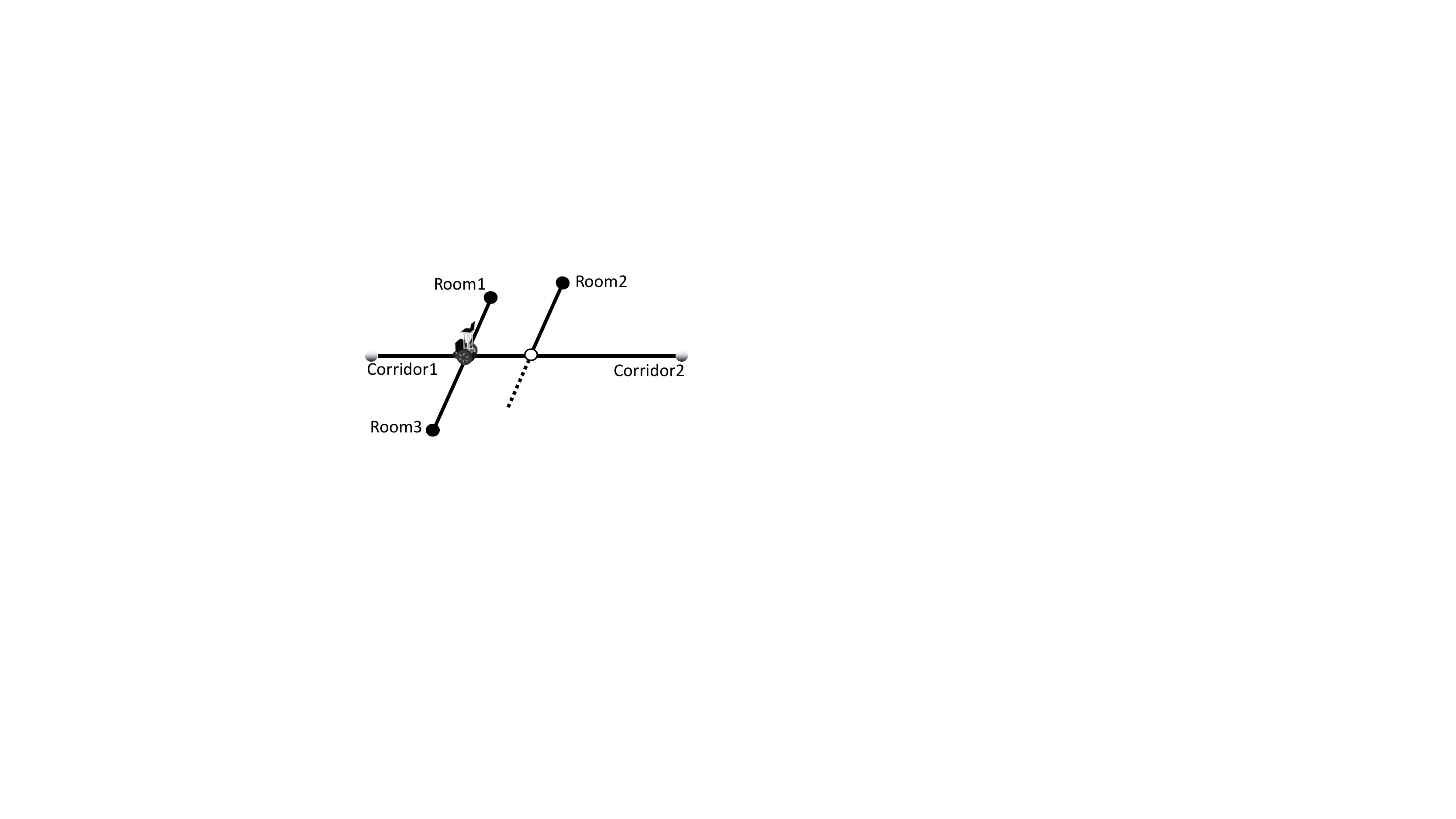}
\caption{ Decision-making with SRM. The semantic information of the node is used for high-level decision-making.}
\label{fig:decisionmaking}
\end{figure}

The target locations during the exploration are always encoded by the nodes on the graph $\mathcal{G}$, hence we can easily query a path to the target on the graph $\mathcal{G}$. The NBT is determined by the high level semantic information, the path cost and the information gain. 

\begin{figure*}[t]
\centering
\begin{minipage}{0.245\textwidth}
\centering
\includegraphics[width=\textwidth]{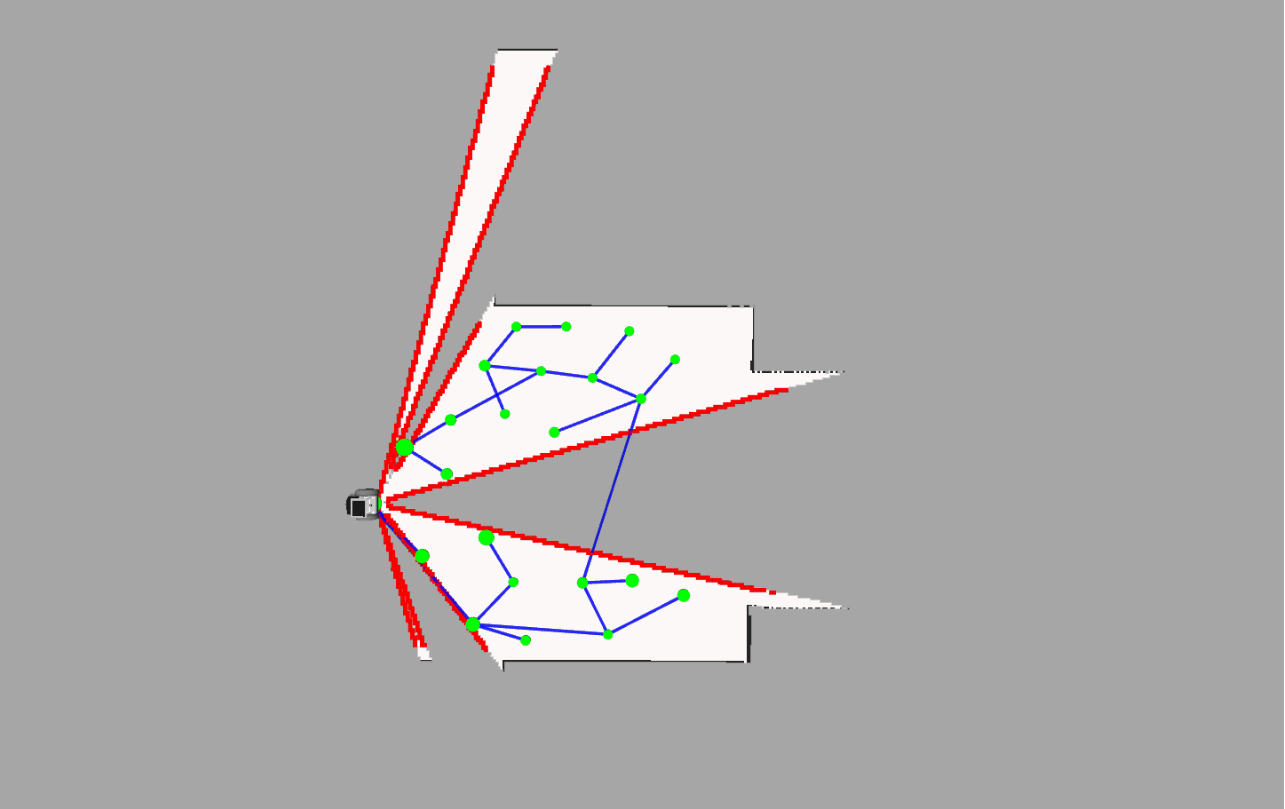}
{(a) T=1s}
\end{minipage} 
\begin{minipage}{0.245\textwidth}
\centering
\includegraphics[width=\textwidth]{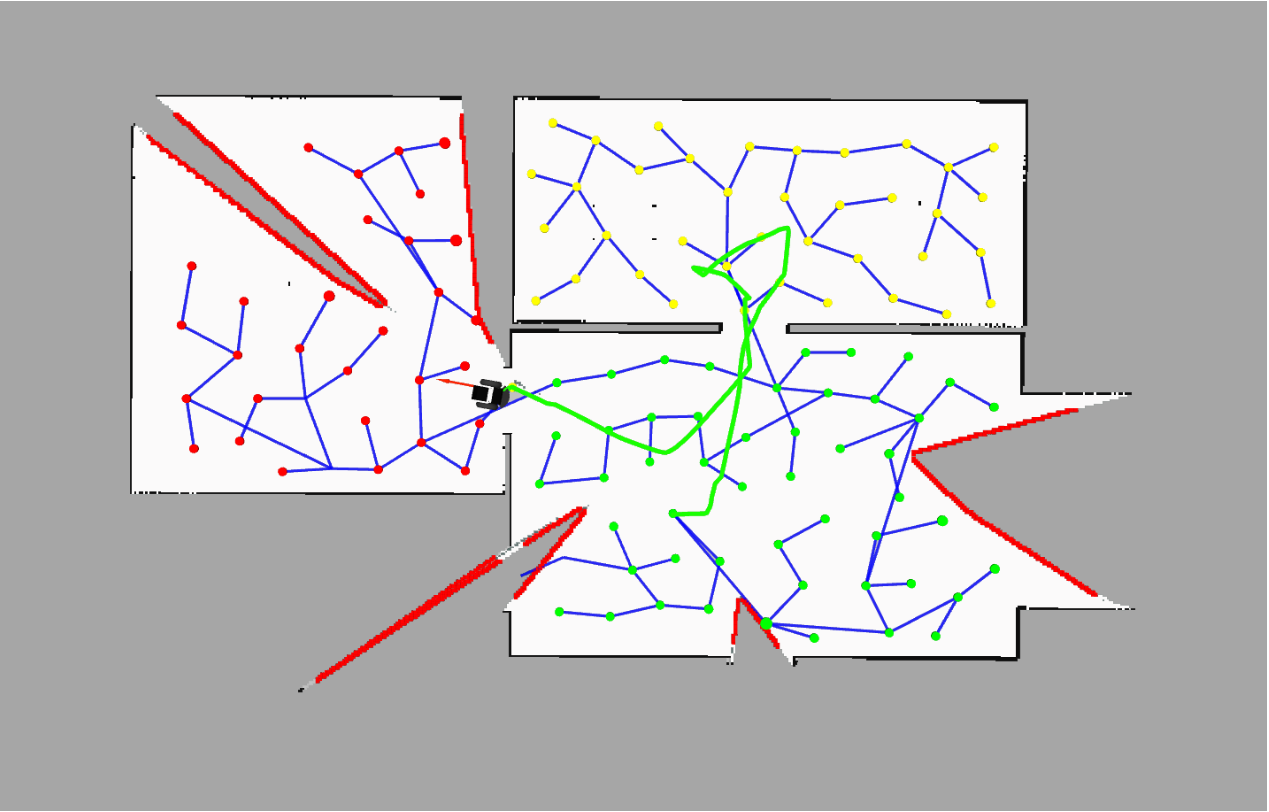}
{(b) T=32s}
\end{minipage} 
\begin{minipage}{0.245\textwidth}
\centering
\includegraphics[width=\textwidth]{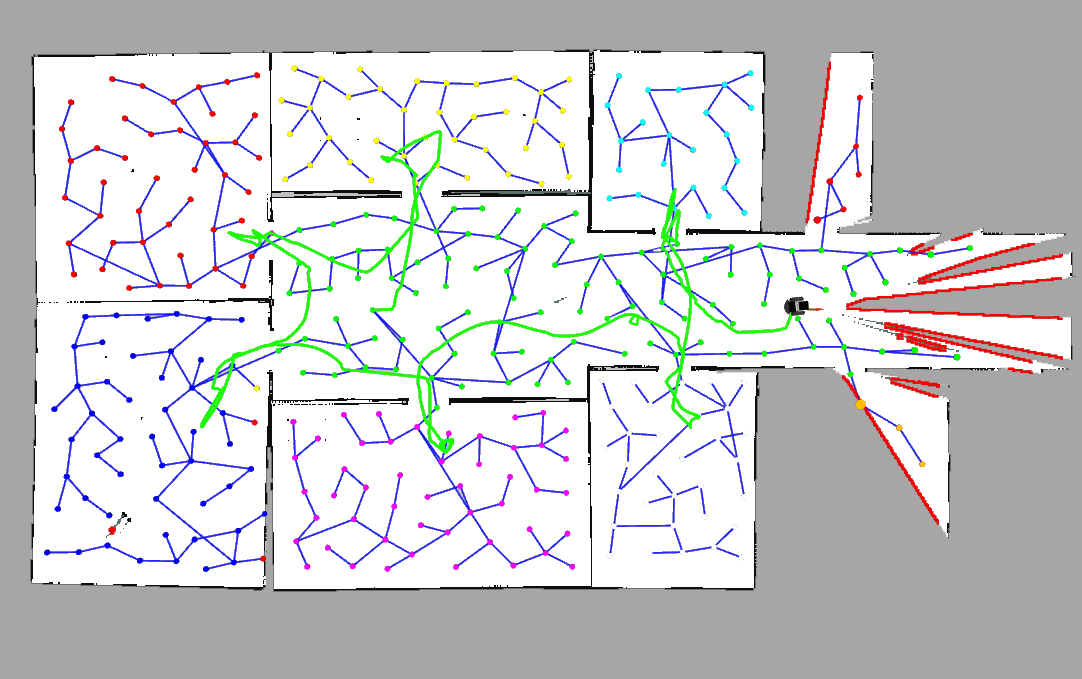}
{(c) T=180s}
\end{minipage}
\begin{minipage}{0.245\textwidth}
\centering
\includegraphics[width=\textwidth]{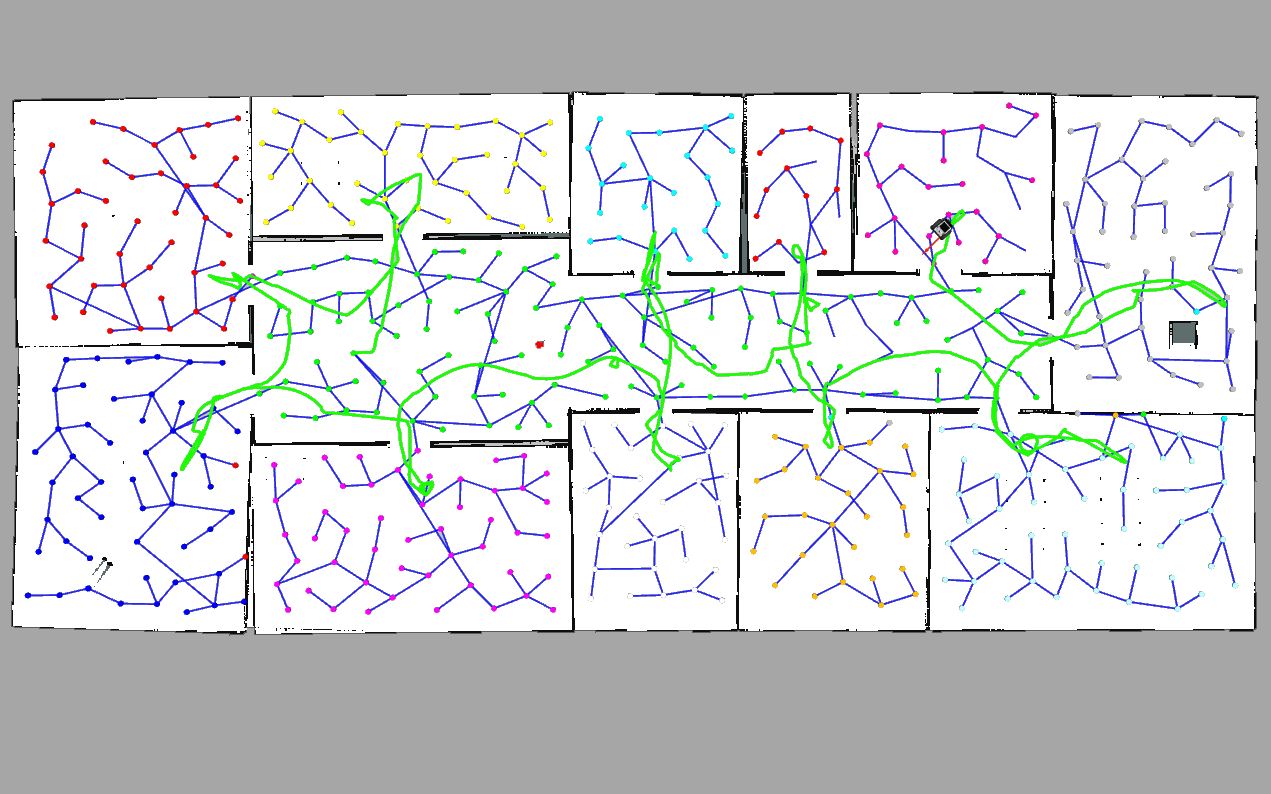}
{(d) T=230s}
\end{minipage}
\vspace{-1mm}
\caption{ Autonomous exploration with SRM in simulated environment at different points of time. The SRM is represented by the colourful nodes and the blue edges. The red stripe indicates the frontiers.}
\label{fig: simlateExperiment}
\vspace{-0.25cm}
\end{figure*} 

The semantic information can give the robot high-level information for guiding the exploration. We divide the scene into two categories, the room $R$ and the connection $C$. $C$ is the place that connects multiple rooms, for example, the corridor. During the exploration, robot always tends to explore the rooms instead of the corridors. As Fig. \ref{fig:decisionmaking} shows, the robot firstly explores the Room 1, Room 2, and Room 3 first according to the distance, then the Corridor 1 and Corridor 2 according to the distance and the information gain. Noticed that when there is no corridor, the exploration will become a Travel Salesman Problem (TSP). Robot will traverse all the rooms according to the effective methods of TSP. Overall the decision is made differently in two cases:
\begin{itemize}
\item \textbf{Case 1:} If there are rooms detected in the environment, the robot will traverse these rooms according to the distance from its current location. When a  new room comes up during this process, the robot will replan this traverse path every time it finishes exploring one room. 

\item \textbf{Case 2:} If there are only corridors in the environment, the robot will select one target according to the reward function $C'=-I(\mathcal{P}_{opt})e^{-\lambda \ell(\mathcal{P}_{opt})}$. Robot considers both the path cost $\ell(\mathcal{P}_{opt})$ and the information gain $I(\mathcal{P}_{opt})$ when selecting the target regions in this case. 
\end{itemize}

Robot will make decisions according to the two cases above. In this simple but effective decision model, robot can distinguish the rooms from the connection places. Generally speaking, the robot is able to explore the room one by one without wasting much time in the corridor. This effectively help reduce the backtracking problem, in which the robot will come back and forth to explore unfinished previous tasks.

\section{Experiments and Results}
\label{sec:experiments and results}

To demonstrate the benefits of our proposed exploration framework, we conduct a number of  experiments in both simulated and real-world environments. The experimental studies demonstrate the effectiveness of our proposed pipeline as well as the efficiency of each individual module of the proposed framework.

We test our proposed framework in a typical simulated indoor environment. As Fig. \ref{fig: simlateExperiment} shows, the blue line indicates the edges and the colorful points indicate the nodes of the graph. Different colors of the node represent different semantic labels. The frontier region, as marked by the red stripe in the Fig. \ref{fig: simlateExperiment}(b)-Fig. \ref{fig: simlateExperiment}(c), can be completely and efficiently detected using the generated SRM. 
The reason is that a lot of nodes that are near the frontiers are generated during SRM buliding process. This means the frontier's location information is already encoded by these nodes, which makes the frontier detection task easier when performing the ray-casting method. This experiment demonstrates that our proposed topological map, which is built incrementally along with the exploration process, can be used for fast frontier detection and guiding the exploration.

\begin{figure}[t]
\centering
\includegraphics[width=7cm,height=5.1cm]{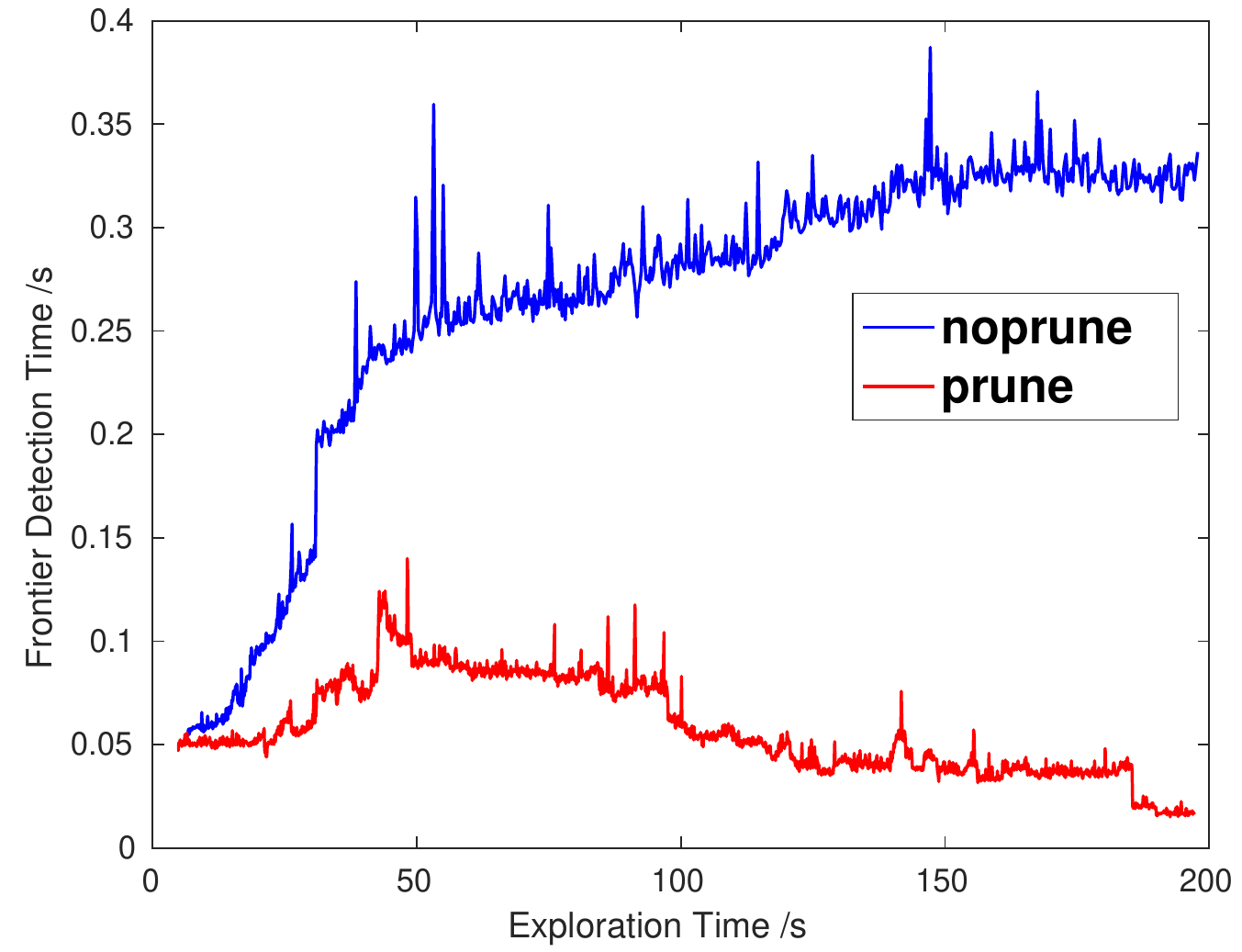}
\caption{Frontier detection time using our proposed method with and without pruning.}
\label{fig:prune}
\end{figure}


\begin{figure*}[t]
\centering
\begin{minipage}{0.245\textwidth}
\centering
\includegraphics[width=\textwidth]{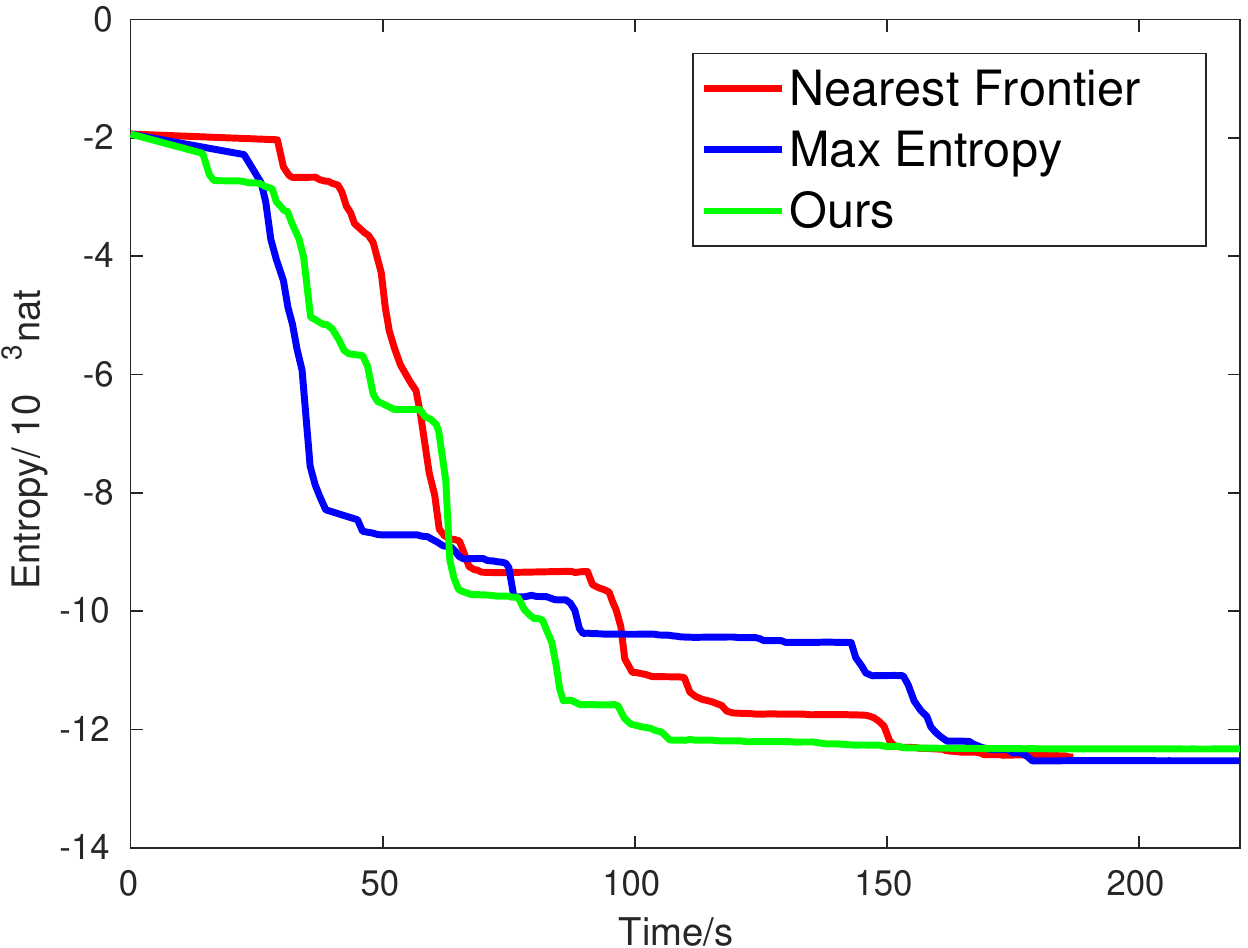}
{Map 1}
\end{minipage} 
\begin{minipage}{0.245\textwidth}
\centering
\includegraphics[width=\textwidth]{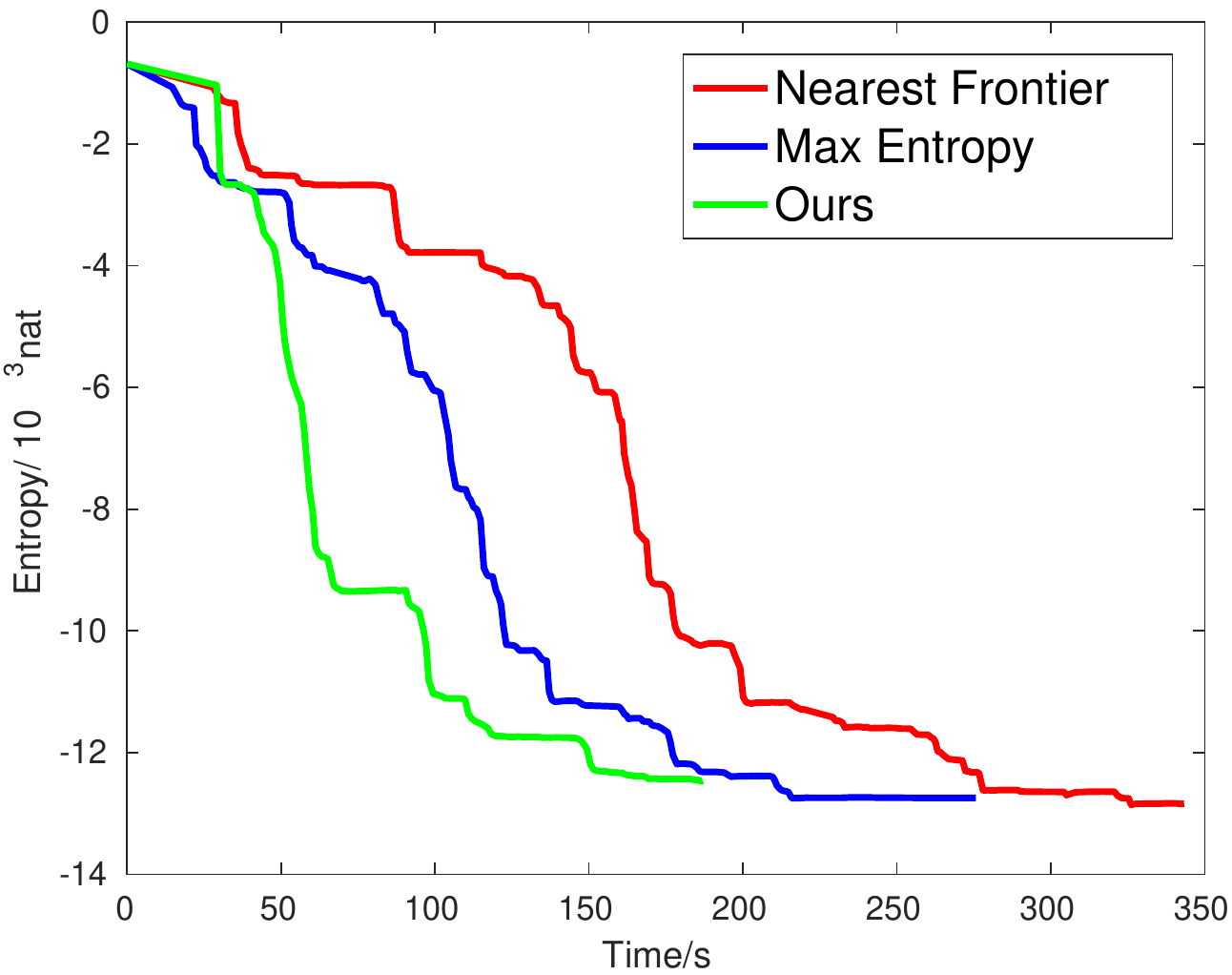}
{Map 2}
\end{minipage} 
\begin{minipage}{0.245\textwidth}
\centering
\includegraphics[width=\textwidth]{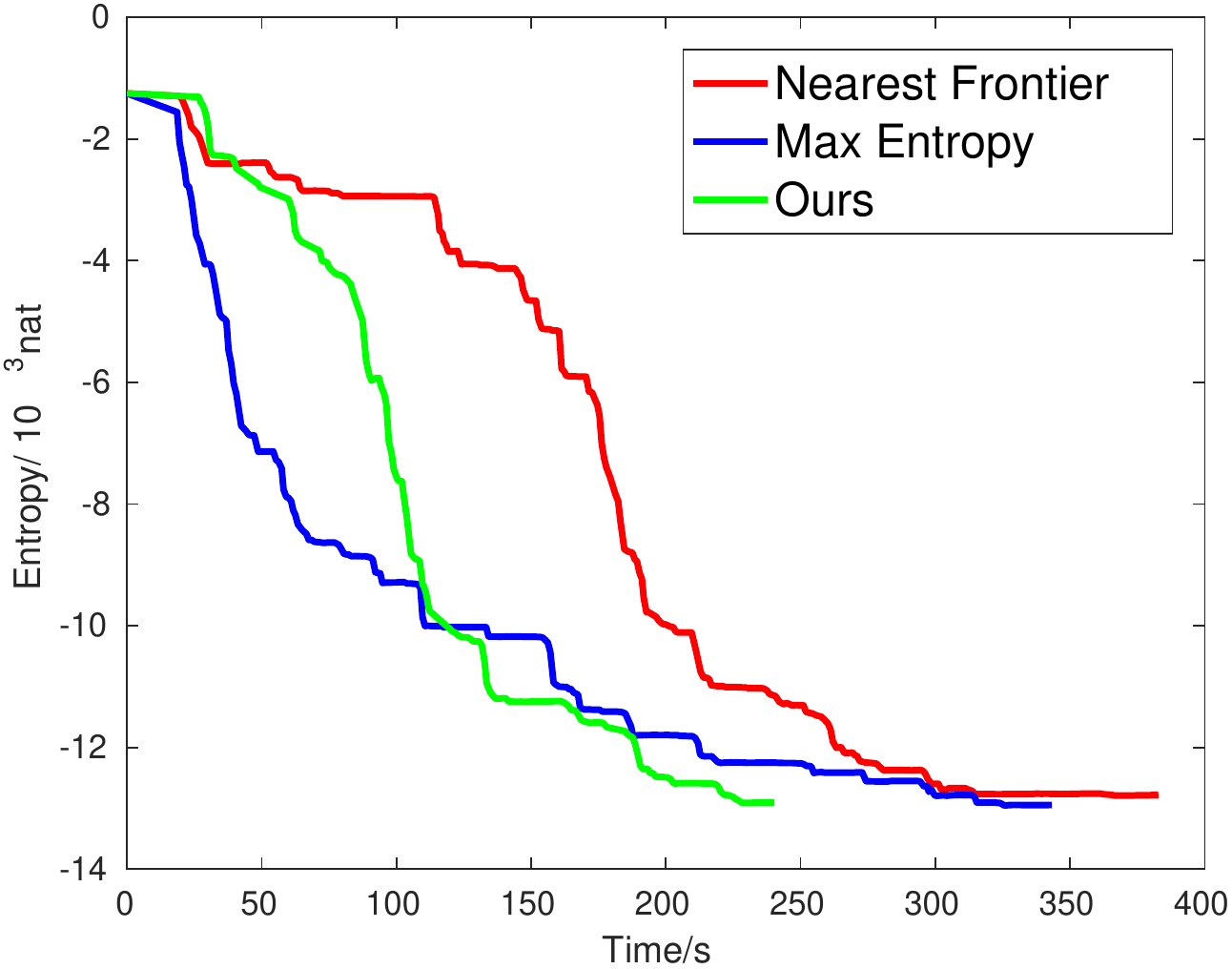}
{Map 3}
\end{minipage}
\begin{minipage}{0.245\textwidth}
\centering
\includegraphics[width=\textwidth]{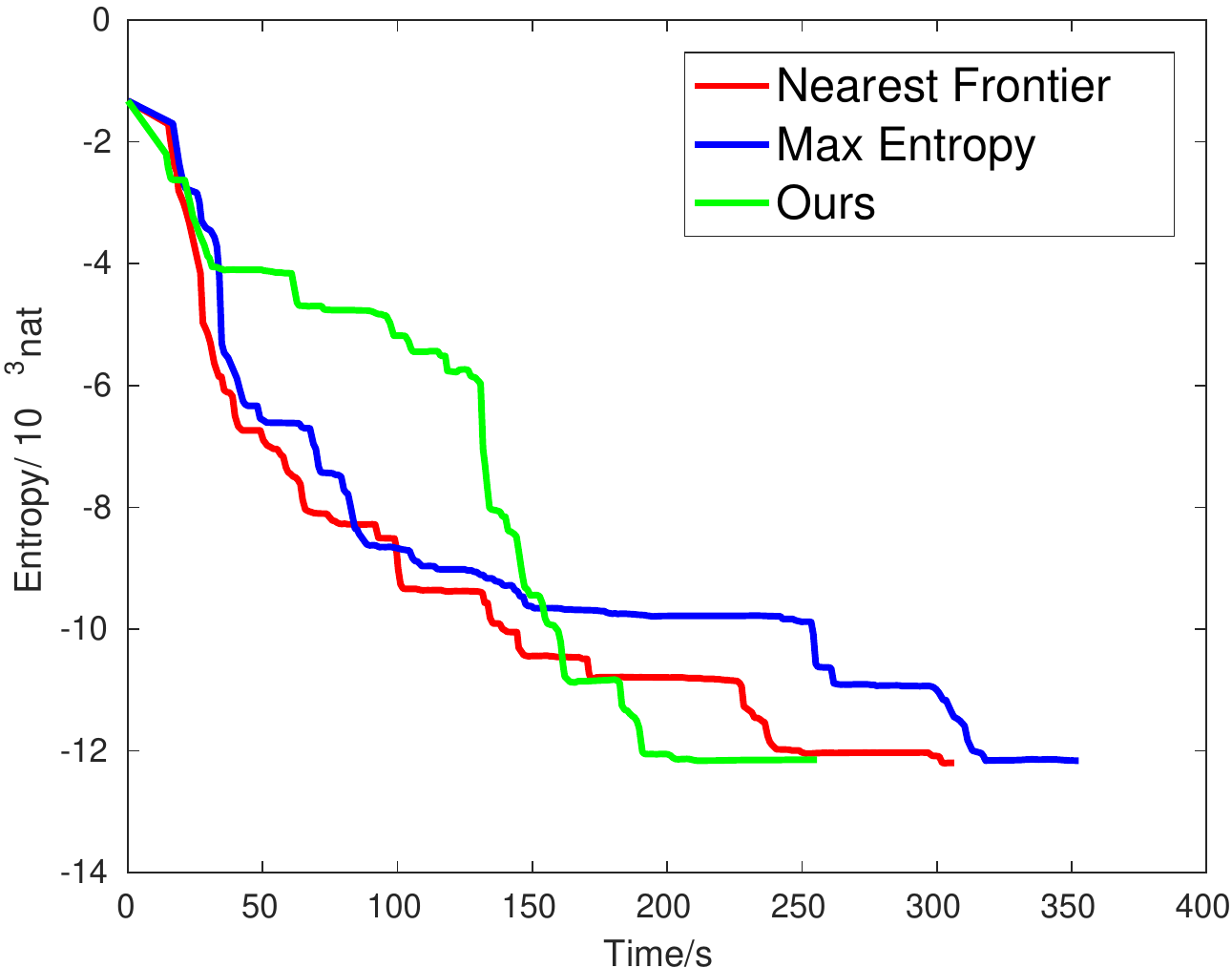}
{Map 4}
\end{minipage}
\vspace{-1mm}
\caption{Variations of map entropy along with the exploration in four different environment scenarios using our proposed method and two comparative methods.}
\label{fig:entropy}
\vspace{-0.25cm}
\end{figure*}  

\begin{figure}[t]
\centering
\begin{minipage}{0.25\textwidth}
\centering
\includegraphics[width=\textwidth,height=3cm]{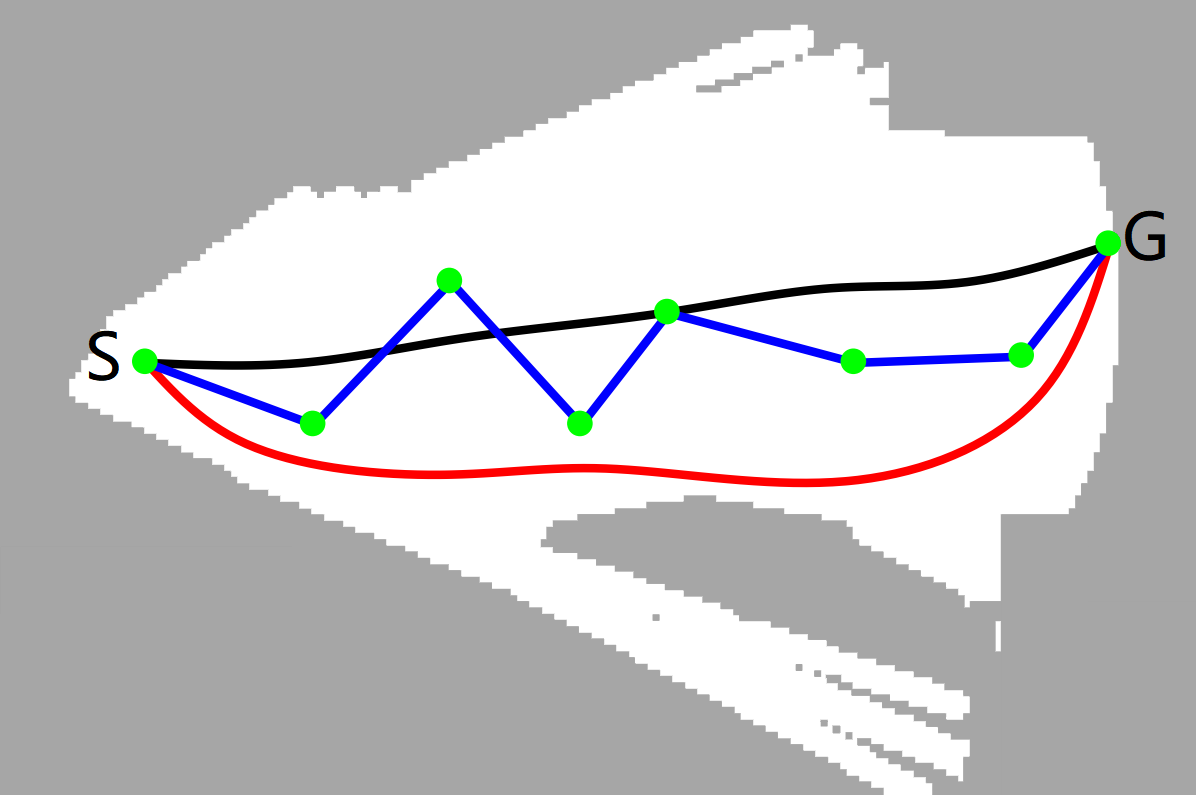}
{(a)}
\end{minipage} 
\begin{minipage}{0.2\textwidth}
\centering
\includegraphics[width=\textwidth]{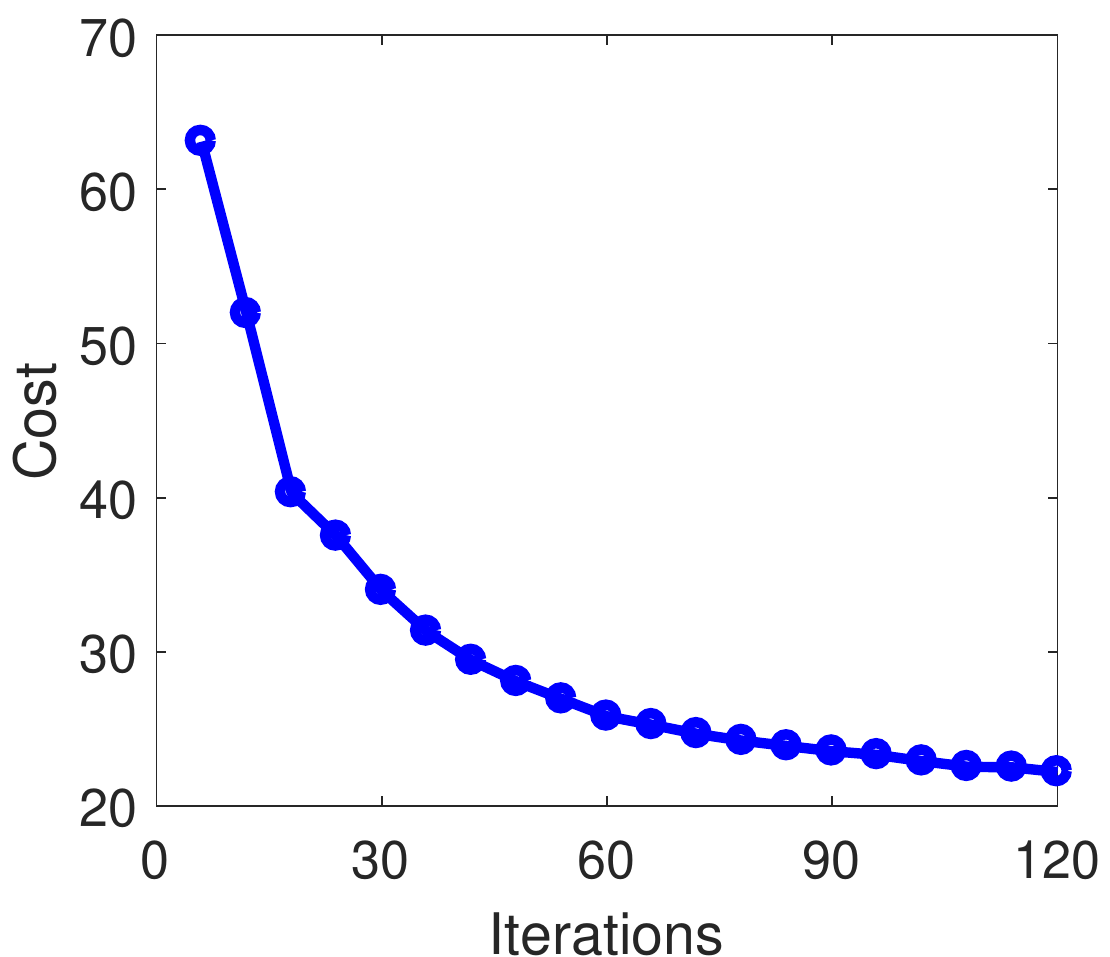}
{(b)}
\end{minipage}
\vspace{-1mm}
\caption{(a) Trajectory optimization using our proposed method. The blue lines are the sparse path queried from the SRM. The black line is the shortest path satisfying the motion constrains. The red line is the informative path generated by our optimization method. (b) The variation of the objective function along with the iterations during the optimization. }
\label{fig:optimize path}
\vspace{-0.25cm}
\end{figure}  

To further explore our frontier detection method, we carried out several experiments. Firstly, we conduct ray-casting on all the nodes of the graph to get the frontier regions. This method is more accurate but it is more time-consuming. Note that there are no frontier regions around some nodes on the graph along with the exploration process, so we propose to prune those nodes.  The experiment is performed in the simulated environment, as shown in Fig. \ref{fig:prune}(a).  Overall, our proposed method can always detect the frontier region within 0.4s, as shown in Fig. \ref{fig:prune}(b). For the method without pruning, the frontier detection time increases significantly along with the exploration, corresponding to the increase of the number of nodes on the graph. The method with pruning is even more efficient after pruning extra nodes. As a quantitative measurement, the red line in Fig. \ref{fig:prune} indicates the frontier detection time after pruning extra nodes. As we can see, the frontier detection time is increasing during the first 50 seconds. This is because there exists a lot of frontier regions and only few extra nodes can be pruned. 
However, the frontier detector can become increasingly efficient with more and more extra nodes being pruned.


\begin{table}[t]
\small
\centering
\caption{Statistics of the comparative study between our proposed planner and the RRT* planner.}
\begin{tabular}{cccccc}
\toprule[1.5pt]
\multirow{2}{*}{Env} &
\multirow{2}{*}{Size ($m^2$)} &
\multicolumn{2}{c}{Our planner} &
\multicolumn{1}{c}{RRT*} \\
\cline{3-6} \\[-0.8em]
 & & Nodes & Time(ms) & Time(ms) &  \\
\midrule
Map1 & 225 & 276  & 0.53ms &7.93ms   \\
Map2 & 300 & 328  & 0.62ms &8.72ms  \\
Map3 & 320 & 359  & 0.72ms &9.37ms  \\
Map4 & 260 & 290  & 0.57ms &8.31ms  \\
\bottomrule[1.5pt]
\end{tabular}
\label{table:planning time}
\end{table}

Robot can query a path efficiently on the topological map $\mathcal{G}$. To demonstrate the efficiency of our proposed planner, we conduct a comparative study in four different environments. The size of each environment is reported in TABLE. \ref{table:planning time}. The control variable is the planner used during the exploration and the nearest frontier based method is adopted for the decision-making module in all the experiments.  We choose RRT* as the comparable planner. As reported in Table. \ref{table:planning time}, our planner is more efficient than the RRT* planner in all the four experimental scenarios.
The time spent on the test group and the control group are both positively related to the map size. Moreover, our method can even better  in large-scale or high-dimensional environments.
 
The performance of our trajectory optimization method is demonstrated in Fig. \ref{fig:optimize path}. The proposed objective function is optimized using the cross-entropy method. As shown in Fig. \ref{fig:optimize path}, the sparse trajectory is queried directly on the generated graph structure. The black line indicates the shortest path satisfying the motion constraints. The useful information is evaluated by the unknown area covered by the robot sensor scope. The black path is short without considering the collected information along the path. Using our proposed cross-entropy based method, we can find a path, as the red line shows in Fig. \ref{fig:optimize path}(a), along which the robot can collect more information. With relatively longer path, the robot can explore more unknown areas when it executes the path from the start to the goal. Fig. \ref{fig:optimize path}(b) shows the variations of the objective function at different steps of the iterations using our cross-entropy based method. Obviously, the value of the objective function is reduced and we can get an optimal path after a few iterations.

 \begin{figure}[t]
 \centering
 \begin{minipage}{0.5\textwidth}
 \centering
 \includegraphics[width=8cm,height=5cm]{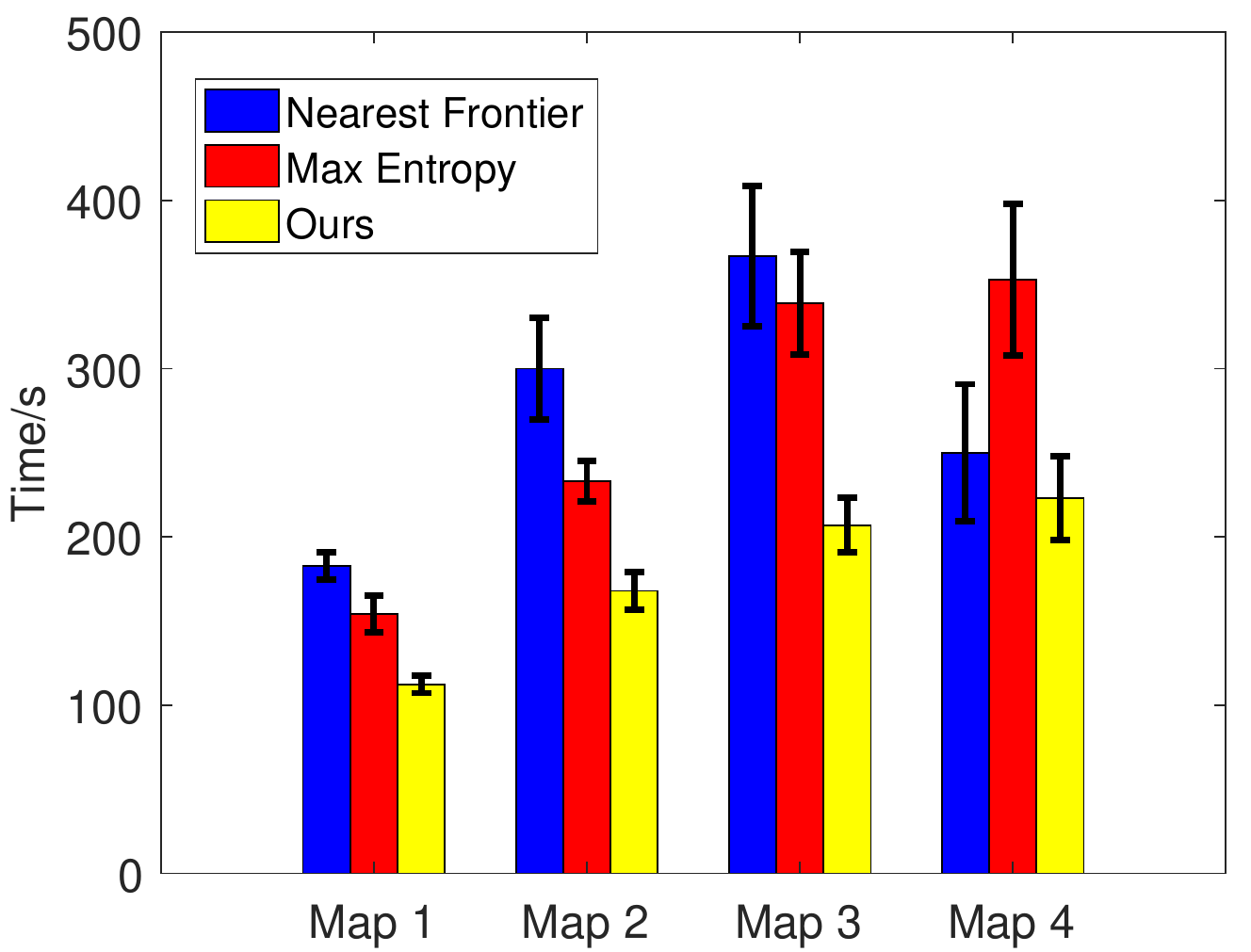}
 \centerline{(a) }
 \end{minipage} 
 \begin{minipage}{0.5\textwidth}
 \centering
\includegraphics[width=8cm,height=5cm]{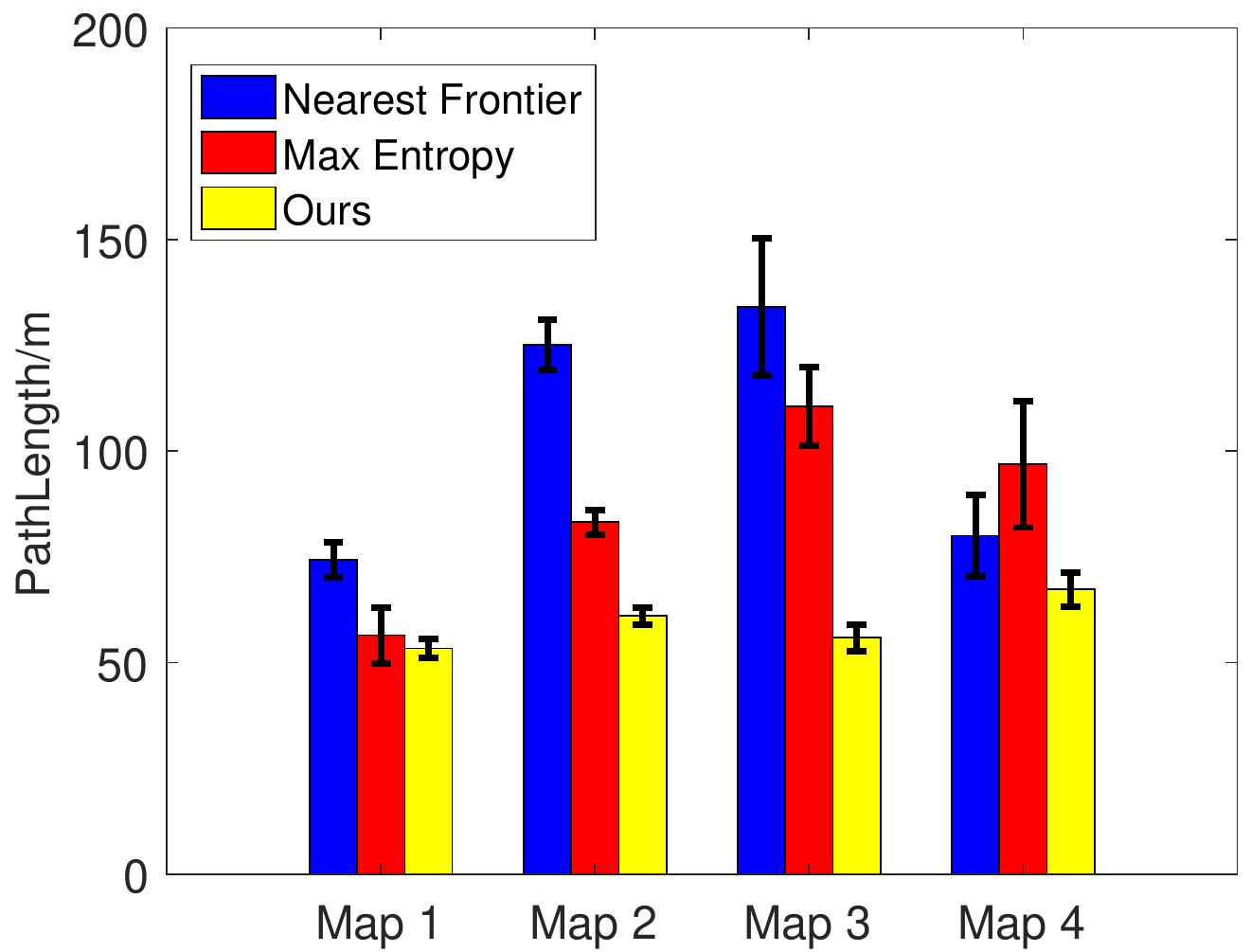}
\centerline {(b) }
 \end{minipage} 
\caption{Exploration time and path cost in four different environments with three methods.}
\label{fig:path length and time spend}
\vspace{-0.25cm}
\end{figure}

We proposed to guide the exploration using the semantic information, the information gain and the path cost to the target. The experiments are performed in four environments to evaluate the efficiency of our exploration strategy. We compare our method with the Nearest frontier based method and the Max Entropy method. The nearest frontier method is a greedy method which always seeks the nearest frontier to explore during the exploration, while the Max Entropy selects the most informative area as the target to explore.  Compared with these two methods, our method is always more efficient in all the four environments, as shown in Fig. \ref{fig:entropy}.  The entropy can be reduced to a smaller value with less time using our method. However, it is difficult to select a better one from the other two methods. The performances of these two methods are highly dependent on the environment. 

The statistics of the path length and time spending are reported in Fig. \ref{fig:path length and time spend}. Our proposed guidance strategy can explore the environment with less time and path cost. Besides, the standard deviation of
our method is also less than the other ones, which means our method works more steadily. There is no much difference between the performance of the Nearest Frontier based method and the Max Entropy method. Furthermore, we can see from Fig. \ref{fig:path length and time spend} that longer path means longer exploration time. This is because of the fact that most exploration time is spent on the execution of the path. The less the path is, the less exploration time is needed. Our proposed guidance strategy is useful in reducing the path cost, as shown in Fig. \ref{fig:path length and time spend}. Besides, the proposed trajectory optimization method can also provide more information with less path cost, which in turn reduce the path length. 

\begin{figure}[!t]
\centering
\includegraphics[width=0.4\textwidth]{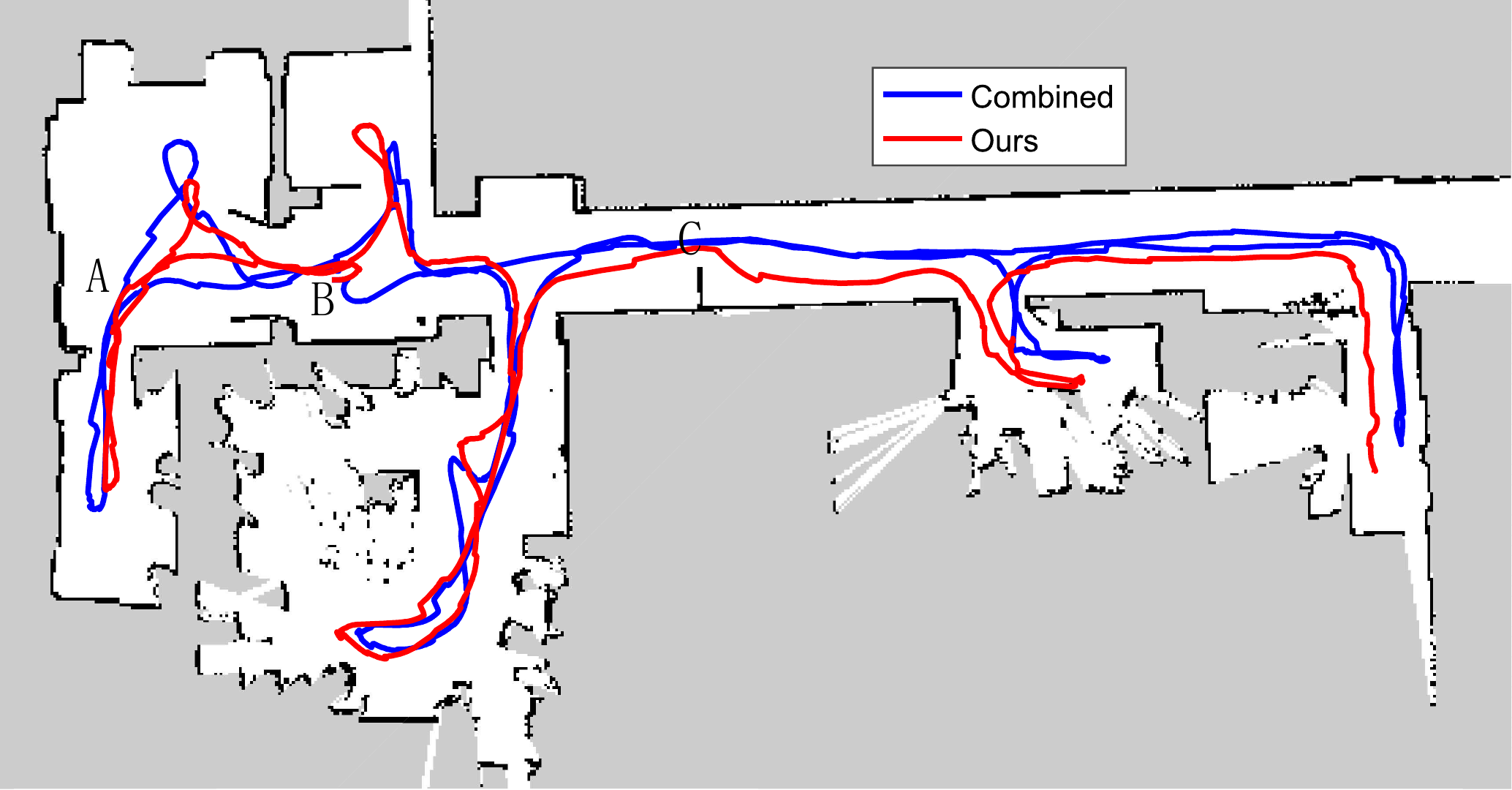}
\vspace{-1mm}
\caption{Real-world experimental study. The blue line and red line represent the exploration trajectory of the combined method (96.5m) and our method (72.2m), respectively.}
\label{fig:real exp}
\vspace{-0.25cm}
\end{figure}

To evaluate the efficiency of our proposed framework, we compare our method with the method that considers both the path cost and the information gain, in which the two criteria are linearly combined weighted by two coefficients $\gamma_1$ and $\gamma_2$.  The combined method uses the RRT* for the path planning during the exploration. Fig. \ref{fig:real exp} shows the experiment environment, where one typical exploration trajectories of our method and the combined method are illustrated. We compare our method with the combined method at three different starting locations and the data is recorded in Table. \ref{table:path cost}. As we can see, the performance of our proposed method is better than the comparison group in all these three scenarios.The combined method may achieve better performance with proper $\gamma_1$ and $\gamma_2$. For example, there is no much difference of the mean path length when starting at location A.  However, the combined method is not stable, which can be seen from the larger standard deviation of the path length.


\begin{table}
\small
\centering
\caption{Comparisons with the combined method at different starting locations.}
\begin{tabular}{cccccc}
\toprule[1.5pt]
\multirow{2}{*}{} &
\multicolumn{2}{c}{Ours} &
\multicolumn{2}{c}{Combined} \\
\cline{2-5} \\[-0.8em]
& Path(m) & Speed(m/s) & Path(m) & Speed(m/s)  \\
\midrule 
A & 72.1 $\pm$ 2.1  & 0.36 & 74.3 $\pm$ 5.2  & 0.30\\
B & 75.8 $\pm$ 4.3  & 0.31 & 98.2 $\pm$ 6.7  & 0.33\\
C & 74.5 $\pm$ 2.7  & 0.27 & 79.7 $\pm$ 5.3  & 0.25\\
\bottomrule[1.5pt]
\end{tabular}
\label{table:path cost}
\end{table}

\section{Conclusion and future work}
\label{sec:conclusion and future work}

In this paper, we propose a novel hybrid map, SRM, for autonomous exploration in indoor environments. This hybrid map features the combination of the semantic map and the topological map. It is a graph structure where a node contains the information gain and the semantic label corresponding to the node location. Both the information gain and the path cost can be obtained efficiently with this graph structure. Moreover, we use the cross-entropy based optimization method to optimize the queried sparse path. This optimization provides a more informative trajectory. The semantic information, combined with the information gain and the path cost, is used to guide the exploration. Our framework is not a simple integration of several modules but a tightly coupled system. The modules in this system contribute to each other to make the exploration more efficient. The experimental studies demonstrate that our method is effective and more efficient than the others.

Our method  currently focuses on the 2D environment. In the future, we will extend our method into 3D exploration area. We plan to use the octomap to represent the environment. Our proposed method can efficiently detect the frontiers and plan a path in 3D environments.

\bibliographystyle{unsrt}
\bibliography{reference}

\begin{thebibliography}{10}

\bibitem{yamauchi1997frontier}
Brian Yamauchi.
\newblock A frontier-based approach for autonomous exploration.
\newblock In {\em Computational Intelligence in Robotics and Automation, 1997.
  CIRA'97., Proceedings., 1997 IEEE International Symposium on}, pages
  146--151. IEEE, 1997.

\bibitem{stachniss2005information}
Cyrill Stachniss, Giorgio Grisetti, and Wolfram Burgard.
\newblock Information gain-based exploration using rao-blackwellized particle
  filters.
\newblock In {\em Robotics: Science and Systems}, volume~2, pages 65--72, 2005.

\bibitem{lavalle1998rapidly}
Steven~M LaValle.
\newblock Rapidly-exploring random trees: A new tool for path planning.
\newblock 1998.

\bibitem{mallios2016toward}
Angelos Mallios, Pere Ridao, David Ribas, Marc Carreras, and Richard Camilli.
\newblock Toward autonomous exploration in confined underwater environments.
\newblock {\em Journal of Field Robotics}, 33(7):994--1012, 2016.

\bibitem{thrun2002robotic}
Sebastian Thrun et~al.
\newblock Robotic mapping: A survey.
\newblock {\em Exploring artificial intelligence in the new millennium},
  1(1-35):1, 2002.

\bibitem{konolige2011navigation}
Kurt Konolige, Eitan Marder-Eppstein, and Bhaskara Marthi.
\newblock Navigation in hybrid metric-topological maps.
\newblock In {\em Robotics and Automation (ICRA), 2011 IEEE International
  Conference on}, pages 3041--3047. IEEE, 2011.

\bibitem{choset2000sensor}
Howie Choset and Joel Burdick.
\newblock Sensor-based exploration: The hierarchical generalized voronoi graph.
\newblock {\em The International Journal of Robotics Research}, 19(2):96--125,
  2000.

\bibitem{rezanejad2015robust}
Morteza Rezanejad, Babak Samari, I~Rekleitis, Kaleem Siddiqi, and Gregory
  Dudek.
\newblock Robust environment mapping using flux skeletons.
\newblock In {\em Intelligent Robots and Systems (IROS), 2015 IEEE/RSJ
  International Conference on}, pages 5700--5705. IEEE, 2015.

\bibitem{keidar2014efficient}
Matan Keidar and Gal~A Kaminka.
\newblock Efficient frontier detection for robot exploration.
\newblock {\em The International Journal of Robotics Research}, 33(2):215--236,
  2014.

\bibitem{jadidi2018gaussian}
Maani~Ghaffari Jadidi, Jaime~Valls Miro, and Gamini Dissanayake.
\newblock Gaussian processes autonomous mapping and exploration for
  range-sensing mobile robots.
\newblock {\em Autonomous Robots}, 42(2):273--290, 2018.

\bibitem{shen2012autonomous}
Shaojie Shen, Nathan Michael, and Vijay Kumar.
\newblock Autonomous indoor 3d exploration with a micro-aerial vehicle.
\newblock In {\em Robotics and Automation (ICRA), 2012 IEEE International
  Conference on}, pages 9--15. IEEE, 2012.

\bibitem{umari2017autonomous}
Hassan Umari and Shayok Mukhopadhyay.
\newblock Autonomous robotic exploration based on multiple rapidly-exploring
  randomized trees.
\newblock In {\em Intelligent Robots and Systems (IROS), 2017 IEEE/RSJ
  International Conference on}, pages 1396--1402. IEEE, 2017.

\bibitem{dornhege2013frontier}
Christian Dornhege and Alexander Kleiner.
\newblock A frontier-void-based approach for autonomous exploration in 3d.
\newblock {\em Advanced Robotics}, 27(6):459--468, 2013.

\bibitem{visser2008balancing}
Arnoud Visser and Bayu~A Slamet.
\newblock Balancing the information gain against the movement cost for
  multi-robot frontier exploration.
\newblock In {\em European Robotics Symposium 2008}, pages 43--52. Springer,
  2008.

\bibitem{osswald2016speeding}
Stefan O{\ss}wald, Maren Bennewitz, Wolfram Burgard, and Cyrill Stachniss.
\newblock Speeding-up robot exploration by exploiting background information.
\newblock {\em IEEE Robotics and Automation Letters}, 1(2):716--723, 2016.

\bibitem{stachniss2009multi}
Cyrill Stachniss.
\newblock Multi-robot exploration using semantic place labels.
\newblock In {\em Robotic Mapping and Exploration}, pages 73--90. Springer,
  2009.

\bibitem{zhu2018rlsuper}
Delong Zhu, Tingguang Li, Danny Ho, and Max Q-H Meng.
\newblock Deep reinforcement learning supervised autonomous exploration in
  office environments.
\newblock In {\em Robotics and Automation (ICRA), 2018 IEEE International
  Conference on}. IEEE, 2018.

\bibitem{heng2015efficient}
Lionel Heng, Alkis Gotovos, Andreas Krause, and Marc Pollefeys.
\newblock Efficient visual exploration and coverage with a micro aerial vehicle
  in unknown environments.
\newblock In {\em ICRA}, volume~3, pages 3--5, 2015.

\bibitem{davis2016c}
Bobby Davis, Ioannis Karamouzas, and Stephen~J Guy.
\newblock C-opt: Coverage-aware trajectory optimization under uncertainty.
\newblock {\em IEEE Robotics and Automation Letters}, 1(2):1020--1027, 2016.

\bibitem{Tan:icra2017}
Y.~T. Tan, Abhinav Kunapareddy, and Marin Kobilarov.
\newblock Gaussian process adaptive sampling using the cross-entropy method for
  environmental sensing and monitoring.
\newblock {\em IEEE International Conference on Robotics and Automation (ICRA),
  Brisbane, Australia}, 2018.

\bibitem{gonzalez2002navigation}
H{\'e}ctor~H Gonz{\'a}lez-Banos and Jean-Claude Latombe.
\newblock Navigation strategies for exploring indoor environments.
\newblock {\em The International Journal of Robotics Research},
  21(10-11):829--848, 2002.

\bibitem{kobilarov2012cross}
Marin Kobilarov.
\newblock Cross-entropy motion planning.
\newblock {\em The International Journal of Robotics Research}, 31(7):855--871,
  2012.

\end{thebibliography}

\end{document}